\documentclass[letterpaper, 10 pt, conference]{ieeeconf}  % Comment this line out if you need a4paper

\IEEEoverridecommandlockouts                              % This command is only needed if 
\usepackage{graphicx} % for pdf, bitmapped graphics files
\usepackage{subcaption} % for pdf, bitmapped graphics files
\usepackage{hyperref}

\usepackage{amsmath} % assumes amsmath package installed
\usepackage{amssymb}  % assumes amsmath package installed

\usepackage{mathtools}
\usepackage{graphicx}
\usepackage{subcaption}
\usepackage{booktabs} % For better table lines
\usepackage{array}    % For better column definitions
\usepackage{color, soul}
\usepackage{algorithm2e}
\usepackage{cite}
\usepackage{svg}
\title{\LARGE \bf

GNN-based Decentralized Perception in Multirobot Systems for Predicting Worker Actions
}

\author{Ali Imran$^{1}$, %Abdalwhab Bakheet Mohamed Abdalwhab$^{1}$, 
Giovanni Beltrame$^{2}$ and David St-Onge$^{1}$ % <-this % stops a space
\thanks{*This work was supported by NSERC and CoRoM.} % <-this % stops a space
\thanks{$^{1}$ are with the Lab of INIT Robots, Ecole de technologie supérieure, 1100 Notre-Dame W., Canada
        {\tt\small ali.imran.1@ens.etsmtl.ca}, {\tt\small david.st-onge@etsmtl.ca}}%
\thanks{$^{2}$ is with MISTLab, Computer and Software Engineering, Polytechnique Montréal, Canada
        {\tt\small giovanni.beltrame@polymtl.ca} 
        }%
\thanks{Project website: https://initrobots.ca/spwi/
}
}%

\newcolumntype{P}[1]{>{\centering\arraybackslash}p{#1}}
\newcolumntype{R}[1]{>{\raggedleft\arraybackslash}p{#1}}

\begin{document}

\maketitle
\thispagestyle{empty}
\pagestyle{empty}

%%%%%%%%%%%%%%%%%%%%%%%%%%%%%%%%%%%%%%%%%%%%%%%%%%%%%%%%%%%%%%%%%%%%%%%%%%%%%%%%
\begin{abstract}
In industrial environments, predicting human actions is essential for ensuring safe and effective collaboration between humans and robots. This paper introduces a perception framework that enables mobile robots to understand and share information about human actions in a decentralized way. The framework first allows each robot to build a spatial graph representing its surroundings, which it then shares with other robots. This shared spatial data is combined with temporal information to track human behavior over time. A swarm-inspired decision-making process is used to ensure all robots agree on a unified interpretation of the human’s actions. Results show that adding more robots and incorporating longer time sequences improve prediction accuracy. Additionally, the consensus mechanism increases system resilience, making the multi-robot setup more reliable in dynamic industrial settings.

\end{abstract}

%%%%%%%%%%%%%%%%%%%%%%%%%%%%%%%%%%%%%%%%%%%%%%%%%%%%%%%%%%%%%%%%%%%%%%%%%%%%%%%%
\section{Introduction}

Collaborative robots are poised to become a cornerstone of Industry 5.0~\cite{xu2021industry}, emphasizing human-centric design solutions to meet the flexibility demands of hyper-customized industrial processes~\cite{maddikunta2022industry}. Significant efforts have been directed toward identifying key enabling technologies to enhance robotic systems with advanced situational awareness and robust safety features for human coworkers. Two pivotal technologies stand out: individualized human-machine interaction systems that merge the strengths of humans and machines, and the application of AI to improve workplace safety~\cite{muller2020enabling}.%\cite{keshvarparast2024collaborative} has categorized human-cobot collaboration into four different types based on their interaction modes: independent collaboration, sequential collaboration, simultaneous collaboration, and supportive collaboration.  

In highly collaborative and hazardous scenarios such as manufacturing facilities, robots must develop a holistic understanding of their environment to ensure efficiency and safety. While humans excel at anticipating events due to their spatial reasoning, understanding of others' behavior, and planning under uncertainty~\cite{thorpe2003field}, robotic systems lack such innate capabilities. However, they can leverage distributed sensing networks and algorithms to model spatial and temporal relationships within a scene.

\begin{figure}[h]
    \centering
    % Top single image spanning the full column width
    \begin{minipage}{\columnwidth}
        \centering
        \includegraphics[width=\textwidth]{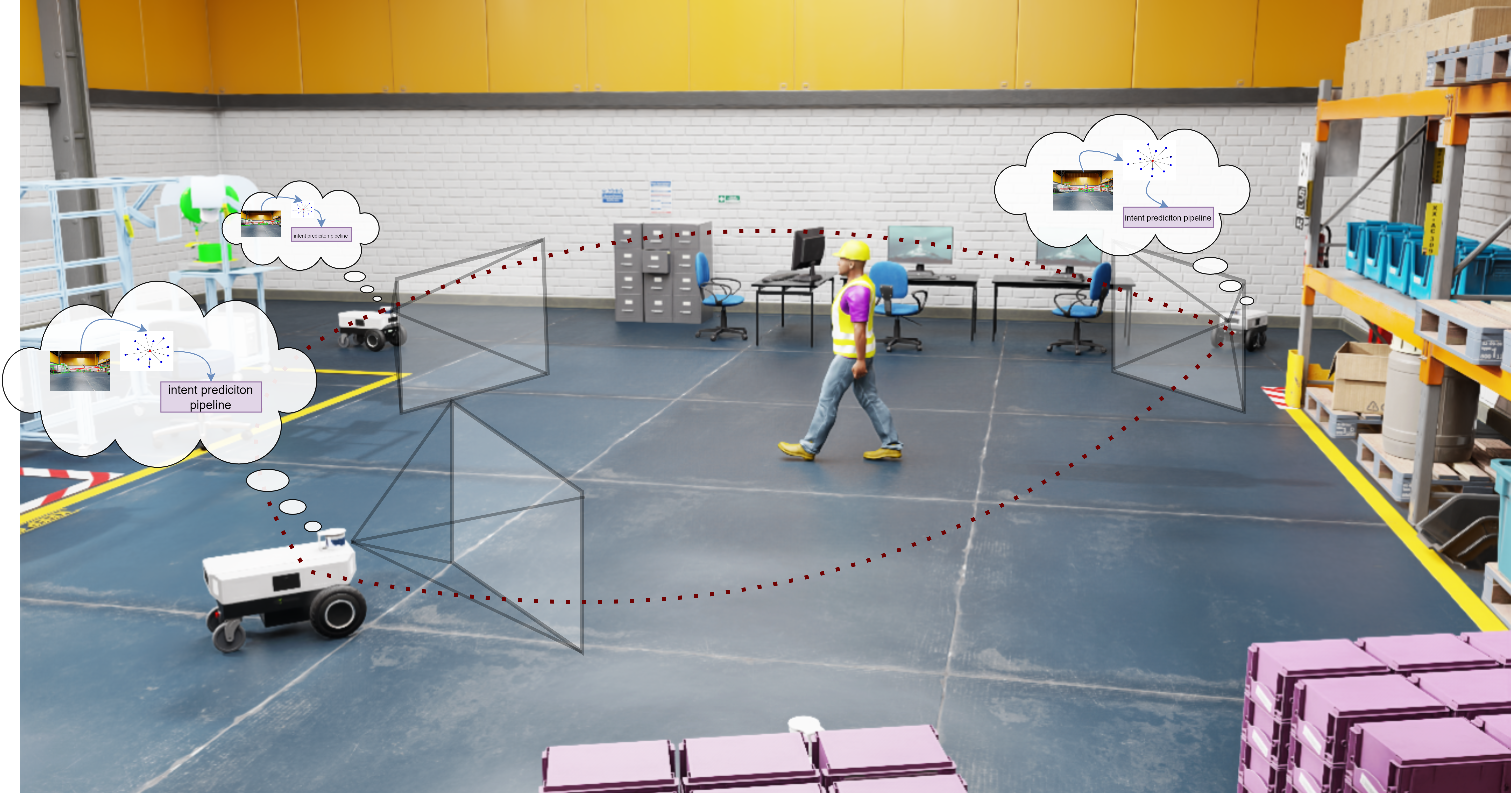}
    \end{minipage}
    
    % Space between images (optional)
    \vspace{0.2cm}
    
    % Three images in a row, each taking 1/3 of the column width
    \begin{minipage}{0.33\columnwidth}
        \centering
        \includegraphics[width=\textwidth]{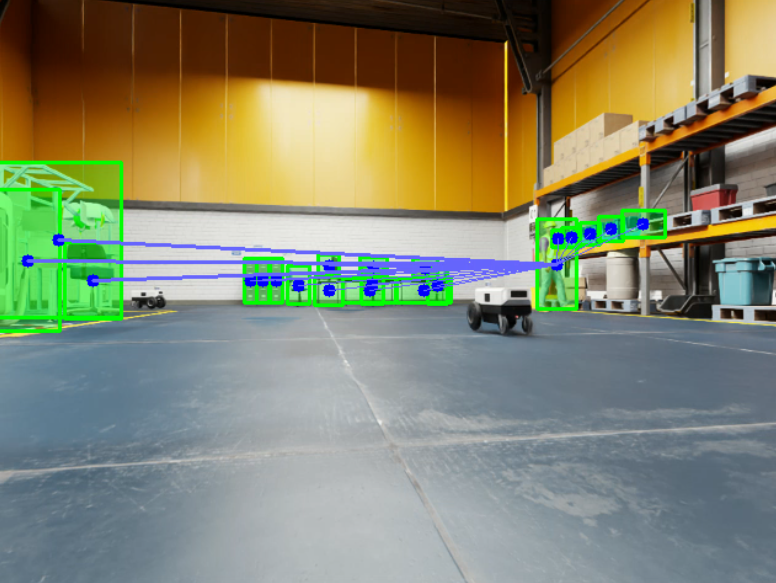}
    \end{minipage}%
    \begin{minipage}{0.33\columnwidth}
        \centering
        \includegraphics[width=\textwidth]{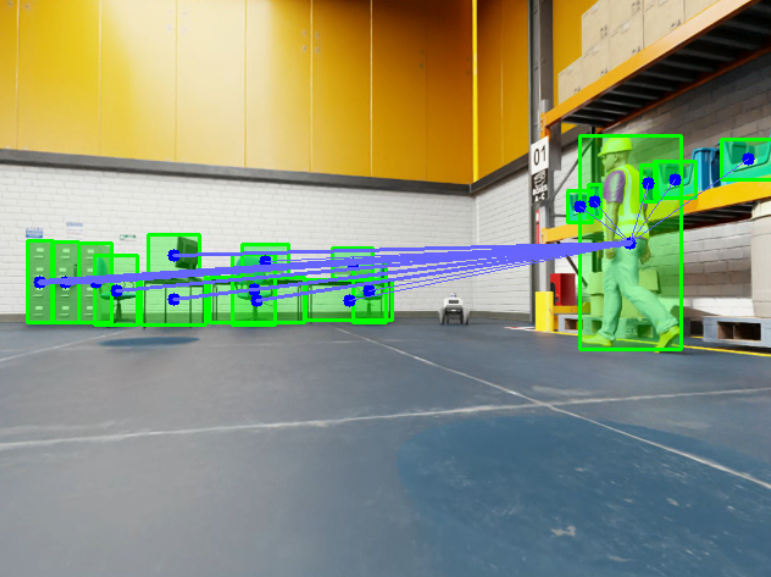}
    \end{minipage}%
    \begin{minipage}{0.33\columnwidth}
        \centering
        \includegraphics[width=\textwidth]{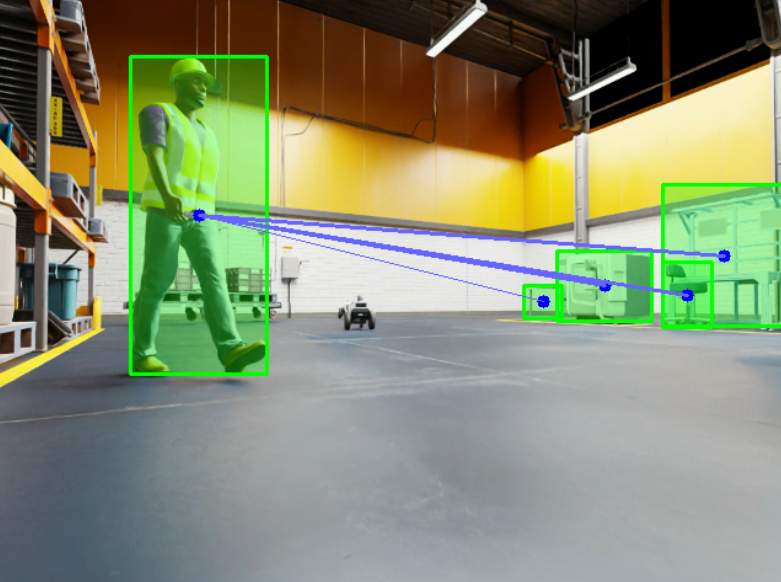}
    \end{minipage}

  \caption{Overview of the multi-robot system deployed in Isaac Sim for human intent prediction. Robots generate spatial graph representations, integrate neighbor information, employ RNNs for temporal understanding, and converge using a swarm-intelligence-inspired consensus mechanism}
  \label{fig:intro}
\end{figure}

\begin{figure*}[!t  ]
  \centering
  \includegraphics[width=\textwidth]{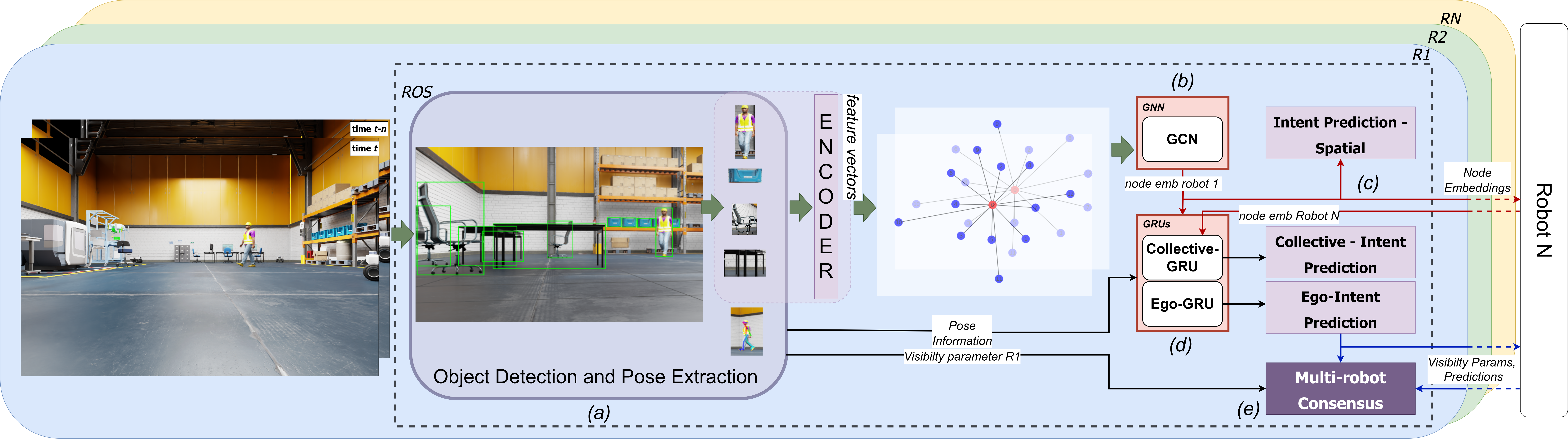}
  \caption{System architecture for multi-robot human intent prediction: a) robot detects the human and objects of interest in the scene, feature vectors are created through an encoder. b) Graph representation of the scene is created. c) Intent prediction based on GNN is made. This information is shared with other robots and the Temporal block. d) Human pose keypoints and spatial information from other robots are aggregated and passed through the temporal understanding block to make ego and collective intent predictions. This information is again shared with other robots. e)  Concensus mechanism utilizes the predictions and quality of visual information from the robot to converge to a single decision about the intent.}
  % Robots capture images, process spatial data with Graph Neural Networks (GNNs), share embeddings, use RNNs for temporal predictions, and converge on a single prediction through a consensus mechanism.
  \label{fig:archi}
\end{figure*}

Algorithms for predicting human behavior and future actions have been explored across domains such as pedestrian crossing prediction~\cite{zhang2017towards, liu2019high, gilroy2023replacing}, intent interpretation in assistive robotics~\cite{gordon2023adaptive, vasquez2013human}, and forecasting human poses in collaborative workspaces~\cite{terreran2020low, bonci2021human}. Most approaches rely on human detection and tracking, using local features like pose, velocity, and location~\cite{quintero2014pedestrian, li2021autonomous, wang2022pedestrian, karasev2016intent}. However, these methods often overlook the relationships between humans and surrounding objects, leading to poor performance when encountering unfamiliar human behavior and limiting their ability to anticipate further into the future.

Recent advancements in graph-based models show good potential for spatiotemporal intent prediction~\cite{liu2020spatiotemporal}. They demonstrate the potential of graph formalisms to emphasize relationships between entities rather than their individual properties, offering generalizable insights~\cite{hamilton2020graph}. In parallel, Graph Neural Networks (GNNs) have gained prominence in applications such as multi-robot path planning~\cite{li2020graph}, collaborative perception~\cite{zhou2022multi}, and navigation in complex environments~\cite{ji2021decentralized}. Building on this foundation, we extend GNN-based approaches to scenarios with a broader range of human actions, leveraging multi-robot systems to develop a shared understanding of the environment.

Multi-robot systems provide significant advantages over single-robot systems, including enhanced robustness, scalability, and flexibility~\cite{parker2016multiple}. While many implementations rely on centralized control, a decentralized approach offers additional benefits, such as improved system flexibility and fault tolerance~\cite{schranz2020swarm}. Decentralization, however, presents its own challenges, particularly in achieving effective coordination and collective decision-making among robots~\cite{brambilla2013swarm}. These challenges arise because, in the absence of a central controller, robots must rely on distributed algorithms to interpret their environment and align their actions.

Consensus mechanisms in swarm intelligence have been widely explored to address these issues. In decentralized multi-robot systems, such as those used in swarm robotics, consensus ensures that robots can collectively make coherent decisions even with diverse inputs and local observations. In decentralized multi-robot systems, such as swarms of robots, consensus can be achieved using strategies like majority voting~\cite{valentini2014self}, ranked voting systems~\cite{shan2021discrete}, and entropy-based local negotiation~\cite{zheng2023consensus}. These methods enable efficient decision alignment while accommodating sensor variability and differences in information quality.

Our shared perception intent prediction pipeline employs graph-based methods to facilitate information exchange between robots. Implemented in ROS, the pipeline integrates data from multiple robots to model spatial relationships between humans and nearby objects using GNNs and temporal relations with Recurrent Neural Networks (RNNs), enabling accurate human intent prediction. This multi-robot strategy enhances robustness by compensating for individual sensor failures and ensuring safety in industrial environments. 
% Additionally, accurate intent prediction would support downstream tasks, such as robot navigation and multirobot task allocation, eventually improving workflows in an industrial setup.

The main contributions of this paper are as follows:
\begin{itemize}
    \item Development of a spatial understanding module leveraging graph neural networks to model relationships between humans and nearby objects.
    \item Introduction of a temporal understanding module that aggregates spatial data from other robots in a decentralized manner to predict and forecast human actions.
    \item Integration of a swarm intelligence-inspired consensus mechanism to ensure decision convergence across all robots in the system.
\end{itemize}

\section{Method}
\subsection{Problem statement}
In manufacturing setups, numerous dynamic processes such as machining, assembly, and material handling occur simultaneously. Human workers perform multiple tasks like machine tending and assembly, requiring adaptability. These dynamic conditions pose significant challenges for robots, which must operate efficiently while prioritizing human safety.

Although manufacturing represents just one subset of potential robot deployment scenarios, it serves as a baseline and motivation for designing our multi-robot system. We define intent prediction as the anticipation of a human operator's future action within a specific time horizon. For this study, we consider time horizons of 1s, 2s, and 3s into the future, based on the \textit{Stopping time and distance metric} outlined in ISO 10218-1:2011 safety standards, Annex B(normative)~\cite{ISO10218_1}. This metric relates to the time required for a robot to detect a human, initiate deceleration, and come to a complete stop, which establishes the minimum threshold for the perception pipeline’s prediction capabilities. Given the relatively low speeds of industrial mobile robots\footnote{MiR250~\cite{mir250} and Otto 1500~\cite{otto1500} both have a top speed of 2.0 m/s}, the selected time horizons are deemed sufficient for safety and operational efficiency. For instance, if a human is detected at a distance of 8 meters while both the robot and the human are moving towards each other, the robot has less than 2 seconds to react\footnote{Assuming a combined closing speed of 4.1 m/s (robot at 2.5 m/s and human at 1.6 m/s), the time to collision is approximately 1.95 seconds when starting 8 meters apart.}. Instead of performing an emergency stop, our system can utilize intent prediction to smoothly adjust the robot's path and avoid potential collisions by predicting into the future.

To evaluate our approach, we modeled a small manufacturing facility where multiple operations occur simultaneously. A human operator attends to various tasks while robots navigate the environment. The operator does not follow predefined or straight paths due to the dynamic nature of the tasks and the need to maneuver around moving robots. This unpredictability, coupled with robot mobility, increases the complexity of perception and intent prediction.

The modeled environment includes four key stations, each representing distinct human actions: 
\begin{enumerate}
    \item \textbf{Storage Area}: Retrieving items from storage shelves.
    \item \textbf{Workstation}: Performing tasks at a workstation.
    \item \textbf{Assembly Station}: Assembling components.
    \item \textbf{Manufacturing Station}: Operating manufacturing equipment.
\end{enumerate}

In this setup, the intent prediction problem consists in predicting the workers movement toward one of these stations. By analyzing the human’s trajectory and interaction with the environment, the model infers which station the human is heading toward, thereby determining their intended action.

% \subsection{Overview}
\subsection{System Overview}
Figure~\ref{fig:archi} shows an overview of the system architecture. 
%We have deployed four mobile robots in a simulated environment, depicted in Figure~\ref{fig:intro}. %Although the robots are equipped with additional sensors beyond RGB cameras, we rely solely on the cameras for this experiment.
%Each robot captures images at regular intervals and publishes them as ROS topics.
The system is activated only when a human is detected in the scene. The image processing pipeline extracts objects from the scene, constructs graph structures, and quantifies the robot's visibility (Section~\ref{img_proc}). These graph structures are then processed through a Graph Neural Network (GNN) (Section~\ref{GNN_stuff}), generating a prediction for the current time step based on spatial understanding. Additionally, node embeddings from the final GNN layer are extracted and shared with other robots in the system.  These embeddings are combined with human pose keypoints from an off-the-shelf object detection algorithm over time to form a sequence. This sequence is passed through a Recurrent Neural Network (RNN), chosen based on whether an ego-centric or collective prediction is required (see Section~\ref{GRU_stuff}). The RNN outputs predictions for the current time step and forecasts actions for future timesteps. The predicted output for the current timestep and a confidence value from the calibrated RNN are shared with other robots. This shared information is then processed within a swarm intelligence-inspired consensus mechanism (see Section~\ref{swarmy_stuff}), where weighted calculations enable the multi-robot system to converge on a unified prediction.

\begin{table}[htbp]
\caption{Objects of Interest}
\label{tab:objects_of_interest}
\centering
\begin{tabular}{c l l}
\toprule
\textbf{Serial No.} & \textbf{Category} & \textbf{Objects} \\
\midrule
1 & Storage Area & Crates, Boxes, Pallets \\
2 & Workstation & Desks, Chairs, Storage Drawers, \\
&&Computers \\
3 & Assembly Station & Workbench, Chair \\
4 & Manufacturing Station & CNC Machine, Table \\
\bottomrule
\end{tabular}
\end{table}

\begin{figure*}[htp]
  \centering
  \includegraphics[height=1.1in]{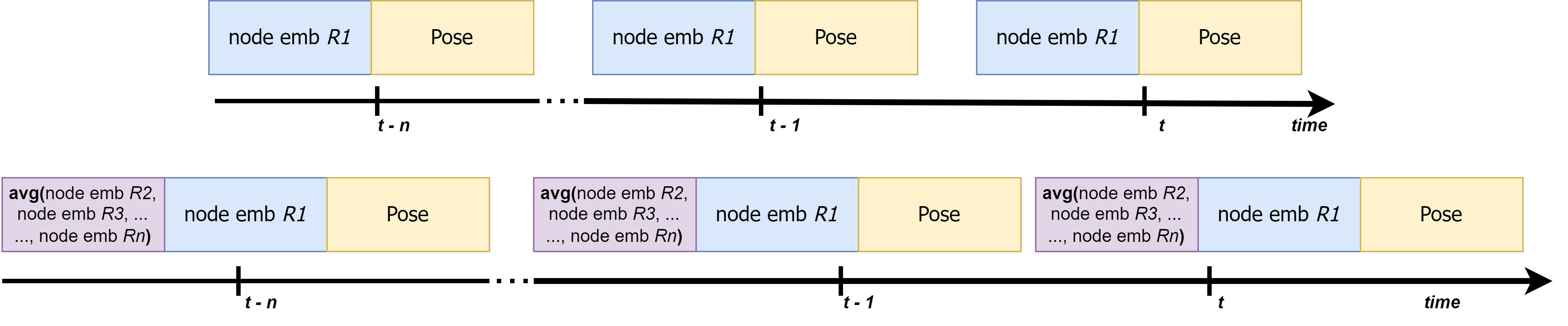}
  \caption{Composition of temporal feature vectors for GRU models. For the Ego-GRU, the feature vector combines the node embeddings from the ego robot and pose information obtained from the object detector. For the Collective-GRU, the feature vector integrates the node embeddings from both the ego robot and neighboring robots, along with the pose information.}
  \label{fig:input}
% \vspace{-5mm}
\end{figure*}

% \subsection{Image processing and graph creation} \label{img_proc} %%%%%%% old
\subsection{Feature Extraction and Representation} \label{img_proc}

% To understand the relationships between the subject and the different objects in the scene, we selected a set of items from the library of preexisting objects commonly found in industrial setups within Isaac Sim. Table~\ref{tab:objects_of_interest} lists these selected objects. In addition to common items like baskets and containers, we included a manufacturing unit and an assembly station to represent specific workstations. While these specific stations may provide valuable insights for better model training, our goal is to observe how well the model can link both specialized equipment and common objects to the human in the scene. To detect both the human and these objects, we employ YOLOv8~\cite{jocher2023yolo}, which has been trained on a custom dataset tailored for our specific human and object detection needs.

We extract feature vectors for every object mentioned in Table~\ref{tab:objects_of_interest} and the human in the scene using a ResNet50~\cite{koonce2021resnet} backbone, which outputs a one-dimensional vector of length 512. For each image, we construct a graph structure where each node represents a concatenation of two feature vectors. The first vector encodes the general appearance of the scene~\cite{zeng2017agent}. The second vector captures the local appearance of specific objects found in the image. Inspired by~\cite{zeng2017agent}, we combine these global and local features by concatenating the two vectors, providing a comprehensive representation for each node in the graph. %effectively 

\subsection{Spatial Relationship}\label{GNN_stuff}

To establish a spatial understanding of the scene, we represent the human and surrounding objects as a star-shaped graph, with the human positioned at the center. Each object is connected to the human node with undirected edges weighted by their Euclidean distances (derived from 2D bounding boxes). These edges form an adjacency matrix $A$, which is augmented with self-loops to create a modified adjacency matrix $A' = A + I$. The node features, initially $H^{(0)}$, are are iteratively transformed through a 2-layer Graph Convolutional Network (GCN)~\cite{kipf2016semi}, leveraging the modified adjacency and degree matrices to aggregate information effectively. 
This shallow 2-layer GCN effectively handles simple star-shaped graphs, minimizing overfitting while ensuring efficient, real-time inference on resource-constrained robotic platforms. Further architectural details are available on the project website.

% \subsection{Temporal GRU}\label{GRU_stuff}
% \subsection{Temporal Understanding}\label{GRU_stuff}
\subsection{Temporal Relationship}\label{GRU_stuff}

To grasp the temporal relationship of the scenario, at least two options have been leveraged in the literature: Gated Recurrent Units (GRUs)~\cite{dey2017gate} and Long Short-Term Memory networks (LSTMs)~\cite{staudemeyer2019understanding}. We select the first due to their lower memory consumption and overall efficiency~\cite{cahuantzi2023comparison}. We have implemented two instances of GRUs: the ego-GRU and the collective-GRU. The ego-GRU processes information from the ego robot without incorporating data from other robots, while the collective-GRU processes aggregated information from multiple robots.

Implementing these two GRUs allows us to gain insights into the multi-robot system. We can evaluate how a single robot performs in understanding the spatial scene and processing temporal information. Additionally, we can assess the improvement in prediction accuracy when a robot incorporates information from other robots, as well as how much it contributes to the overall prediction accuracy of the system.

The input vectors for the ego-GRU and collective-GRU are illustrated in Figure~\ref{fig:input}. For the ego-GRU, we concatenate the output of the final layer of the GNN—which is a 1D vector of length 128—with the flattened output of the human keypoints detector (a 1D vector of length 34). This provides a local, ego-centric spatiotemporal prediction.

% \subsection{Decentralized Information Sharing}\label{zenoh_stuff} %%%% old heading
\subsection{Decentralized Collective Prediction}\label{zenoh_stuff}

% Effective coordination among robots for collective behavior relies heavily on inter-robot communication. As the number of robots increases, managing communication becomes increasingly complex. To address this challenge, we adopted Zenoh~\cite{zenoh}, a publish/subscribe/query protocol that facilitates decentralized communication. Zenoh allows robots to share information efficiently. By implementing publishers and subscribers for each robot's messages through a single topic at the application layer. This approach allows robots to be seamlessly added or removed from the system without disrupting operations. Moreover, Zenoh introduces minimal communication overhead, requiring only 5 bytes per packet.

Messages are used to share node embeddings (outputs of GNNs from other robots) and combined in the collective-GRU model. We use Zenoh~\cite{zenoh}, which allows robots to share information efficiently and also serves as a middleware for ROS 2. We adopt a straightforward aggregation strategy, averaging all the node embeddings to produce a unified representation. This input format is designed to be both adaptable and scalable, as all inputs are reduced to a fixed-sized vector regardless of the number of participating robots. By leveraging this aggregated data, each robot enhances its predictions with information collected from other robots’ perspectives. However, at this stage, achieving convergence to a single, unified decision about human intent remains unresolved.

%%%%%%%%%%%%%%%%%%%%%%%%%%%%
% \RestyleAlgo{ruled}
% \begin{algorithm}[h]
% \caption{Multi-robot Consensus Algorithm}
% \label{alg:consensus_simple}
% \KwIn{From all robots $\{v_i, \hat{p}_i, c_i\}_{i=1}^n$}
% \KwOut{Consensus prediction $\hat{p}^*$, number of participating robots $n$}

% Initialize action set $\mathcal{A} \gets \{0, 1, 2, 3\}$; weights $w(a) \gets 0$, $\forall a \in \mathcal{A}$\;

% Collect $\mathcal{V} \gets \{v_i\}$, $\mathcal{P} \gets \{\hat{p}_i\}$, $\mathcal{C} \gets \{c_i\}$\;

% $n \gets |\mathcal{V}|$\;

% \If{$n > 1$}{
%     $\tilde{\mathcal{V}} \gets \mathrm{Norm}(\mathcal{V})$, $\tilde{\mathcal{C}} \gets \mathrm{Norm}(\mathcal{C})$\;
% }
% \Else{
%     $\tilde{\mathcal{V}} \gets \mathcal{V}$, $\tilde{\mathcal{C}} \gets \mathcal{C}$\;
% }

% \For{each robot $i$}{
%     $w(\hat{p}_i) \mathrel{+}= w_v\,\tilde{v}_i + w_c\,\tilde{c}_i$\;
% }

% $\hat{p}^* \gets \arg\max_{a \in \mathcal{A}} w(a)$\;

% \Return $\hat{p}^*$, $n$\;

% \end{algorithm} 
%%%%%%%%%%%%%%%%%%%%%%%%%%%

\subsection{Consensus Mechanism}\label{swarmy_stuff}

To achieve consensus among multiple robots regarding human intent, we implement a consensus algorithm that aggregates individual predictions, weighting them by each robot’s visibility ratio and prediction confidence. Inspired by majority rule mechanisms in collective decision-making~\cite{montes2011majority, valentini2013majority, scheidler2011dynamics}, this approach is designed to scale with varying numbers of robots.

% We define \(v_i = \tfrac{N_{\text{detected},i}}{N_{\text{total}}}\) and \(c_i = \mathrm{softmax}\bigl(z_{i,k}/T\bigr)\). Letting \(M\) be the number of robots, we then normalize these values as \(\tilde{v}_i = v_i / \sum_{r=1}^M v_r\) and \(\tilde{c}_i = c_i / \sum_{r=1}^M c_r\). A higher visibility ratio indicates a more comprehensive view of the scene by robot $i$. \(z_{i,k}\) is logit the score for the top selected class, and \(T\) tunes the softmax distribution. The weighted vote is thus:
% \[
% V_i = \alpha \tilde{v}_i \;+\; \beta \tilde{c}_i.
% \]

% Here, \(\alpha,\beta\) are scalar weights balancing visibility and confidence. Each robot’s predicted action (from a four-class softmax) and its visibility/confidence values yield a weighted vote. After normalization, we select the action with the highest weighted score and record the number of participating robots. For additional details, please refer to the project website.

We define the visibility ratio for each robot \(v_i = \tfrac{N_{\text{detected},i}}{N_{\text{total}}}\), where \( N_{\text{detected},i} \) is the number of detections by robot \( i \) and \( N_{\text{total}} \) is the total number of detections across all robots. The prediction confidence is defined as \(c_i = \mathrm{softmax}\bigl(z_{i,k}/T\bigr)\), where \( z_{i,k} \) is the logit score for the top selected class, and \( T \) is a temperature parameter that tunes the softmax distribution. Let \( M \) be the number of robots. We then normalize these values as \(\tilde{v}_i = v_i / \sum_{r=1}^M v_r\) and \(\tilde{c}_i = c_i / \sum_{r=1}^M c_r\), where \( v_r \) and \( c_r \) denote the visibility ratio and confidence for each robot \( r \), respectively. The weighted vote for each robot is calculated as:
\[
V_i = \alpha \tilde{v}_i \;+\; \beta \tilde{c}_i.
\]
where \( \alpha \) and \( \beta \) are scalar weights that balance the contributions of visibility and confidence, respectively. A higher visibility ratio \( \tilde{v}_i \) indicates a more comprehensive view of the scene by robot \( i \). Each robot’s predicted action (from a four-class softmax) and its corresponding weighted vote \( V_i \) are aggregated across all robots. For additional details, please refer to the project website.

% \section{Simulation}\label{sec:sim}
\section{Experiments}\label{sec:sim}

We evaluated our approach using a realistic simulation environment provided by NVIDIA's IsaacSim~\cite{nvidia_isaac_sim}. This platform offers a lifelike industrial setting complete with ROS support, human trajectory planning, and synthetic dataset generation capabilities. The simulation environment we created is depicted in Figure~\ref{fig:intro}.

Human motion and actions were simulated using IsaacSim's default custom motion model~\cite{nvidia_isaac_sim}, which incorporates reactive collision avoidance and velocity-based position estimation to navigate both static and dynamic obstacles. During experimentation, the human's goal position for each station remained constant, while the initial position was randomly varied with each simulation run to generalize learning.

We deployed between one and four Carter robots~\cite{nvidia_isaac_sim} from IsaacSim's library. Each robot operated independently, using its onboard sensors to perceive the environment. Robots shared their observations with others in the system, enabling collaborative inference of the human’s intended action.

\subsection{Evaluation Strategy}

We perform simulation-in-the-loop testing, where the starting positions of the robots and the navigation paths differed from those used during training. 
% Additionally, the initial position of the human operator was varied while the destination remained constant.% Since we anticipate future action, the evaluation is done offline.
Ground truth values and predictions were recorded, and the ground truth data was aggregated to match the length of the prediction horizon.

To evaluate the system's ability to infer human actions under varying conditions, we designed a comprehensive strategy incorporating three distinct scenarios, each progressively increasing in difficulty. These scenarios tested the robustness and adaptability of the system in action inference:

\textbf{Scenario 1: Ideal Conditions} In the first scenario, there are no obstacles present in the environment, and the robots remain stationary. This setting allows the human worker to move in a straight path toward the intended station.

\textbf{Scenario 2: Static Obstacles Introduced} In the second scenario, we introduce static obstacles into the environment while keeping the robots stationary. The presence of obstacles forces the human to navigate around them, resulting in less predictable trajectories.

\textbf{Scenario 3: Dynamic Environment with Moving Robots} The third scenario is the most complex and dynamic. In addition to static obstacles, the robots are also moving within the environment. This creates a highly dynamic setting where both the human and the robots are in motion, and the scene changes continuously.

To further challenge the system, two of the stations were placed in close proximity - details of the environment layout are available on the website - testing the system’s ability to distinguish between similar actions when potential destinations were very close together. 
% By evaluating how well the system can differentiate between closely situated goal positions, we assess its precision in action inference under spatial ambiguity.

To validate our approach, we compared its performance against two alternative methods inspired by prior work~\cite{liu2020spatiotemporal}:

\textbf{Constant Velocity Model (CVM)} The CVM is a non-deep learning approach that predicts future states based on simple velocity assumptions. Using an object detection algorithm, we tracked the human's head to extract keypoints and employed a Kalman filter for temporal tracking. The twelve latest states are used to calculate the average velocity and predict the worker's goal.

\textbf{1D Convolutional Neural Network (1D-CNN)} The 1D-CNN served as a baseline to evaluate whether simpler models could suffice without requiring graph neural networks (GNNs). Given the one-dimensional nature of the input data, a 1D-CNN aligns well with the task. We implemented a five-layer 1D-CNN architecture inspired by~\cite{abu2018will}, adapting the input and output to suit our application. The node feature of the human was used as input, and the output was integrated with GRUs to function similarly to our spatiotemporal pipeline.

\begin{figure*}[ht]
    \centering
    \begin{subfigure}{0.32\textwidth}
        \centering
        \includegraphics[scale=0.70]{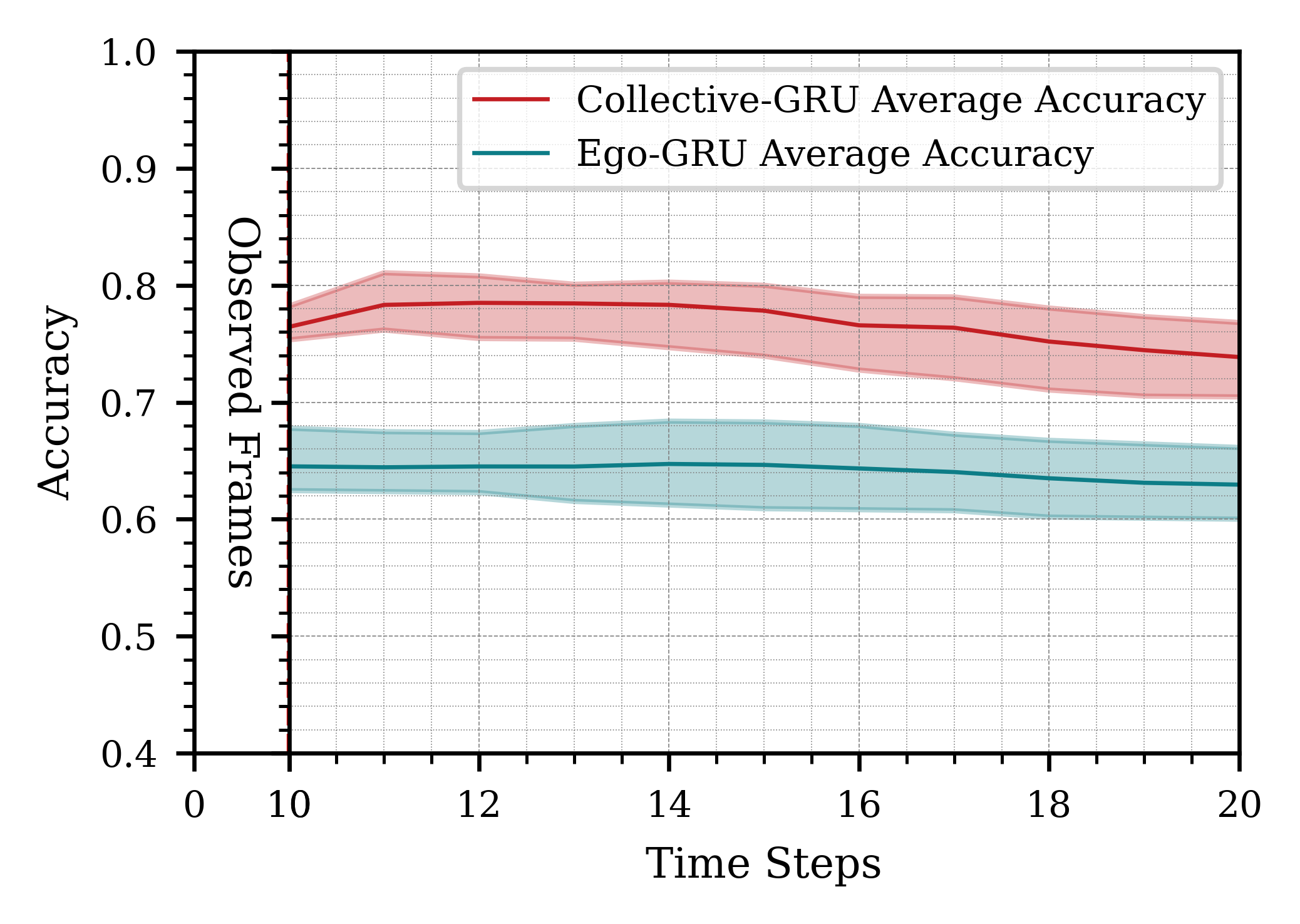}
        \caption{$1s \rightarrow 1s$}
        % \label{fig:subfig1}
    \end{subfigure}
    \hspace{0.0\textwidth} % Reduced spacing
    \begin{subfigure}{0.32\textwidth}
        \centering
        \includegraphics[scale=0.699]{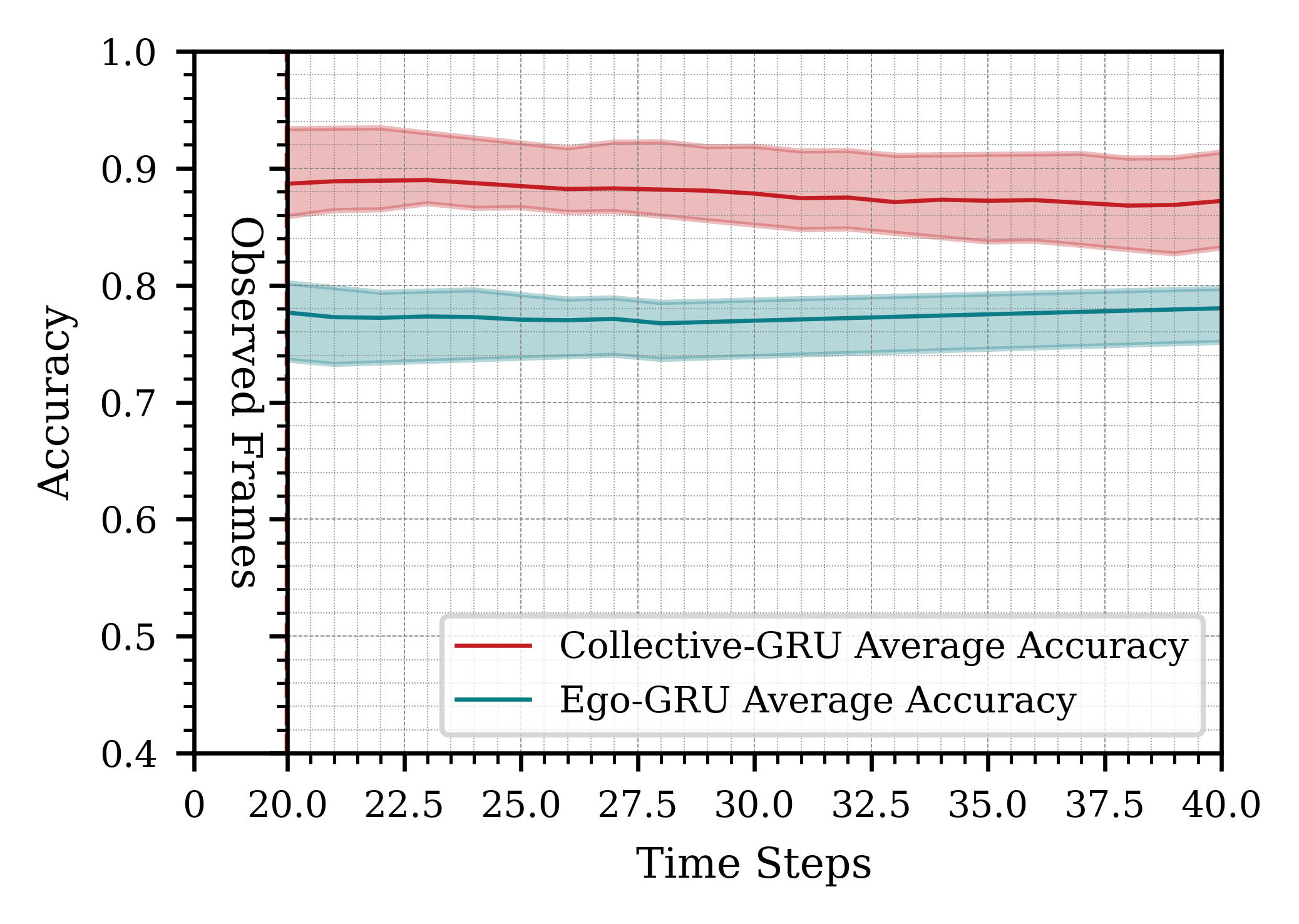}
        \caption{$2s \rightarrow 2s$}
        % \label{fig:subfig2}
    \end{subfigure}
    \hspace{0.0\textwidth} % Reduced spacing
    \begin{subfigure}{0.32\textwidth}
        \centering
        \includegraphics[scale=0.70]{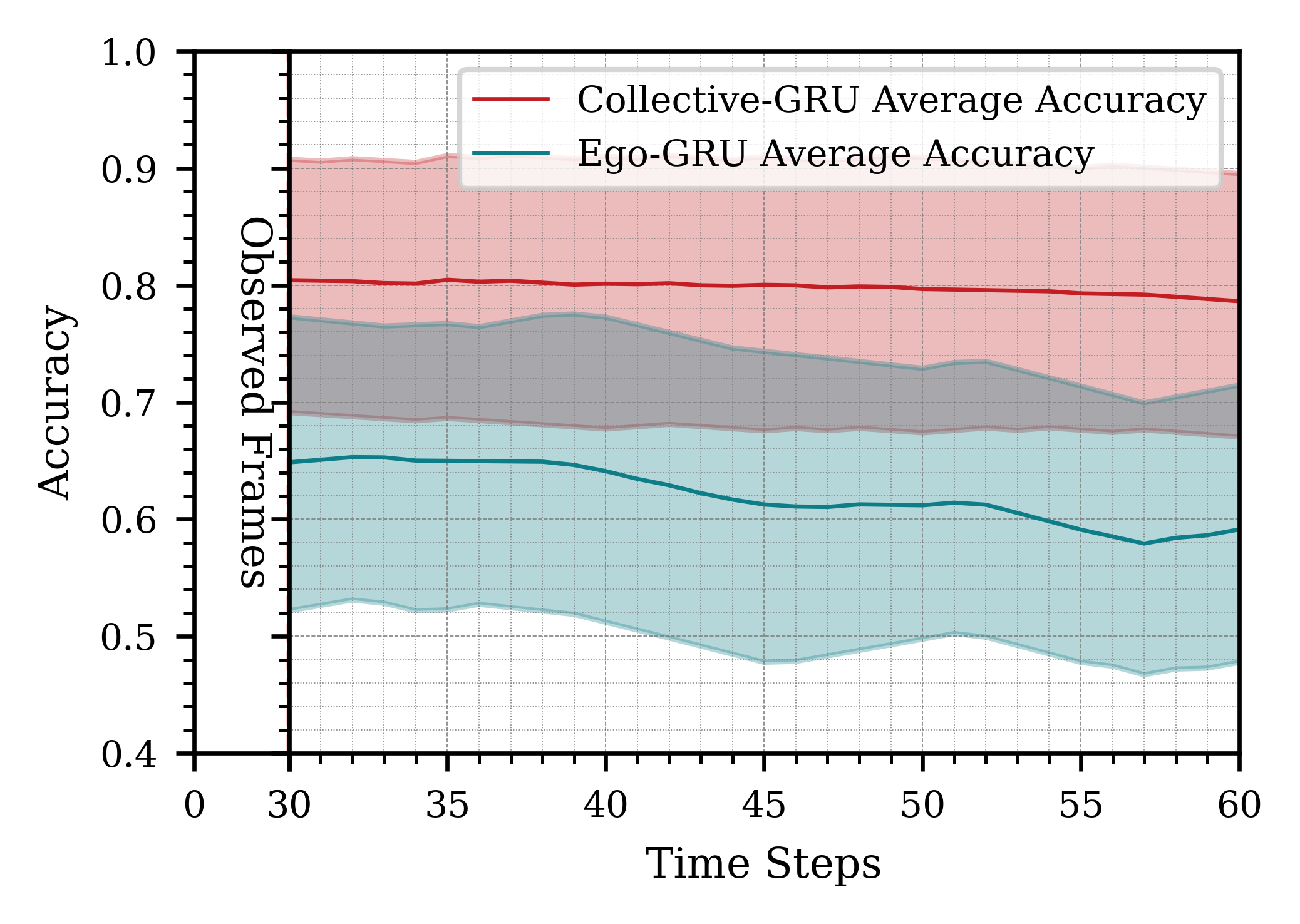}
        \caption{$3s \rightarrow 3s$}
        % \label{fig:subfig3}
    \end{subfigure}
    \caption{Performance comparison of Ego-GRU and Collective-GRU (with 3 robots) models using different temporal horizons: (a) 1s past observation predicting 1s into the future, (b) 2s past predicting 2s into the future, (c) 3s past predicting 3s into the future.}
    \label{fig:horizon}
\end{figure*}

\subsection{Dataset}

We collected three datasets corresponding to three different simulation scenarios. These datasets were designed to capture the motion, appearance, and state of human operator within the scene from the perspectives of multiple mobile robots positioned at various locations. The datasets consist of temporal sequences; each sequence is a set of frames where the human is either: moving towards a goal, transitioning from one goal to another, and stationary. To ensure effective learning, we maintained an equal distribution of all three cases. Moreover, since we are dealing with a multi-robot system, we extended the datasets to include sequences for all possible combinations involving two robots, three robots, and four robots, respectively. A standard split of 60\%/20\%/20\% was implemented for training, validation, and testing.

\begin{figure}[ht]
    \centering
    \includegraphics[width=0.9\linewidth]{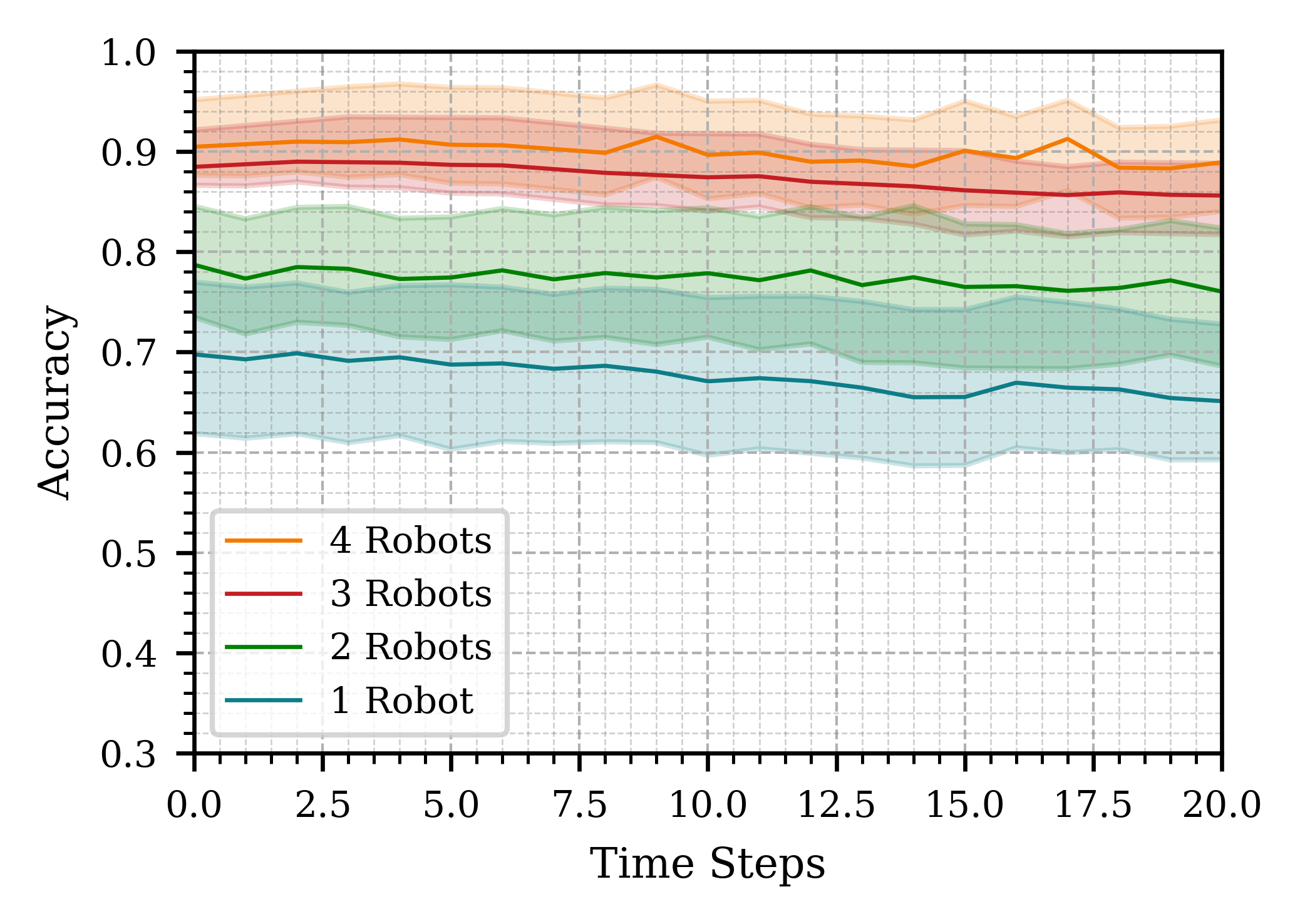}
    \caption{: Effect of increasing the number of robots in our multi-robot system on prediction accuracy.}
    \label{fig:num_robo}
\end{figure}

\section{Results and Discussion}
We present the simulation results for data processed at 10 Hz. Overall, implementing a multi-layered, multi-robot perception strategy significantly enhanced the robustness of predictions. Table~\ref{tab:scenario_accuracy} shows the system's performance across different simulation scenarios described in  Section~\ref{sec:sim}. As expected, the accuracy of the system decreases in more challenging scenarios.

\begin{table}[ht]
\centering
\caption{Average accuracy of the Collective-GRU across three simulation setups of increasing difficulty. Scenario 1: No obstacles and stationary robots, Scenario 2: Obstacles and stationary robots, Scenario 3: Obstacles and moving robots. Metrics include predictions at time step $t$ and average for future timesteps $t+1$ to $n=20$.}
\label{tab:scenario_accuracy}
\small % Reduce font size if necessary
\begin{tabular}{%
    l@{\hspace{4pt}}% Robot names
    R{0.8cm}@{\hspace{4pt}}% GNN
    P{0.8cm}@{\hspace{4pt}}% t under Ego%
    P{1cm}@{\hspace{4pt}}% avg[t+1,n] under Ego%
    P{0.8cm}@{\hspace{4pt}}% t under Collective%
    P{1cm}@{\hspace{4pt}}% avg[t+1,n] under Collective%
    R{1cm}% Consensus%
}
\toprule
\textbf{Scenario} & \textbf{GNN} & \multicolumn{2}{c}{\textbf{Ego\%}} & \multicolumn{2}{c}{\textbf{Collective\%}} & \textbf{Consensus} \\
\cmidrule(lr){3-4} \cmidrule(lr){5-6}
& \% & \textbf{$t$} & \textbf{[$\!1$, $n$]} & \textbf{$t$} & \textbf{[$1$, $n$]} & \% \\
\midrule
Scenario 1 & 71.21 & 84.2 & 80.5 & 92.80 & 90.30 & 90.3 \\
Scenario 2 & 73.40 & 82.45 & 79.82 & 91.83 & 91.02 & 89.33 \\
Scenario 3 & 70.73 & 77.7 & 73.5 & 88.87 & 88.1 & 88.5 \\
% \midrule
% \textbf{Overall} &\textbf{63.04} & \textbf{68.29} & \textbf{66.07} & \textbf{80.89} & \textbf{81.00} & \textbf{80.31} \\
\bottomrule
\end{tabular}
\end{table}

\begin{table}[ht]
\centering
\caption{Breakdown of the multi-robot perception system: enhancements in prediction accuracy, contributions to collective perception, and effectiveness of the consensus mechanism. Metrics include predictions at time step $t$ and average for future timesteps $t+1$ to $n=20$.}
\label{table:robot_accuracy}
\small % Reduce font size if necessary
\begin{tabular}{%
    l@{\hspace{4pt}}% Robot names
    R{0.8cm}@{\hspace{4pt}}% GNN
    P{0.8cm}@{\hspace{4pt}}% t under Ego%
    P{1cm}@{\hspace{4pt}}% avg[t+1,n] under Ego%
    P{0.8cm}@{\hspace{4pt}}% t under Collective%
    P{1cm}@{\hspace{4pt}}% avg[t+1,n] under Collective%
    R{1cm}% Consensus%
}
\toprule
\textbf{Robot} & \textbf{GNN} & \multicolumn{2}{c}{\textbf{Ego\%}} & \multicolumn{2}{c}{\textbf{Collective\%}} & \textbf{Consensus} \\
\cmidrule(lr){3-4} \cmidrule(lr){5-6}
& \% & \textbf{$t$} & \textbf{[$\!1$, $n$]} & \textbf{$t$} & \textbf{[$1$, $n$]} & \% \\
\midrule
Robot 1 & 69.2 & 76.1 & 72.3 & 86.3 & 86.1 & \textbf{88.5} \\
Robot 2 & 72.90 & 79.2 & 75.1 & 91.8 & 90.5 & \textbf{88.5} \\
Robot 3 & 70.1 & 77.8 & 73.10 & 88.5 & 87.8 & \textbf{88.5} \\
\midrule
\textbf{Overall} &\textbf{70.73} & \textbf{77.7} & \textbf{73.5} & \textbf{88.87} & \textbf{88.1} & \textbf{88.5} \\
\bottomrule
\end{tabular}
\end{table}

The GNN models, which rely solely on spatial information, demonstrate consistent accuracy across all scenarios. This consistency indicates that spatial features alone provide a stable but limited understanding of the environment. This can be explained based on the fact that these rely only on the spatial information albeit without leveraging temporal data. Consequently, their performance remains limited to immediate spatial relationships. In contrast, the collective-GRU consistently outperforms the ego-GRU across all scenarios, demonstrating the value of incorporating data from multiple robots.

The consensus mechanism plays a crucial role in enhancing the robustness of the multi-robot system's decisions. By aggregating individual predictions weighted by confidence and visibility, it ensures that the system's overall decision is less affected by individual inaccuracies. While the consensus accuracies are close to the highest individual model accuracies, they remain relatively stable across different scenarios, providing consistent performance.

This trend can also be seen in Table~\ref{table:robot_accuracy}, which provides deeper insights into the contributions of each robot. For instance, in a specific case with three robots, where the system observes 2 seconds of past data and predicts 2 seconds into the future, individual robot accuracies vary. However, the consensus mechanism consistently achieves stable overall accuracy, demonstrating its effectiveness in leveraging collective insights to enhance prediction robustness. This pattern holds across all robots.

\begin{figure}[h]
    \centering
    % Three images in a row, each taking 1/3 of the column width
    \begin{minipage}{0.47\columnwidth}
        \centering
        \includegraphics[width=\textwidth]{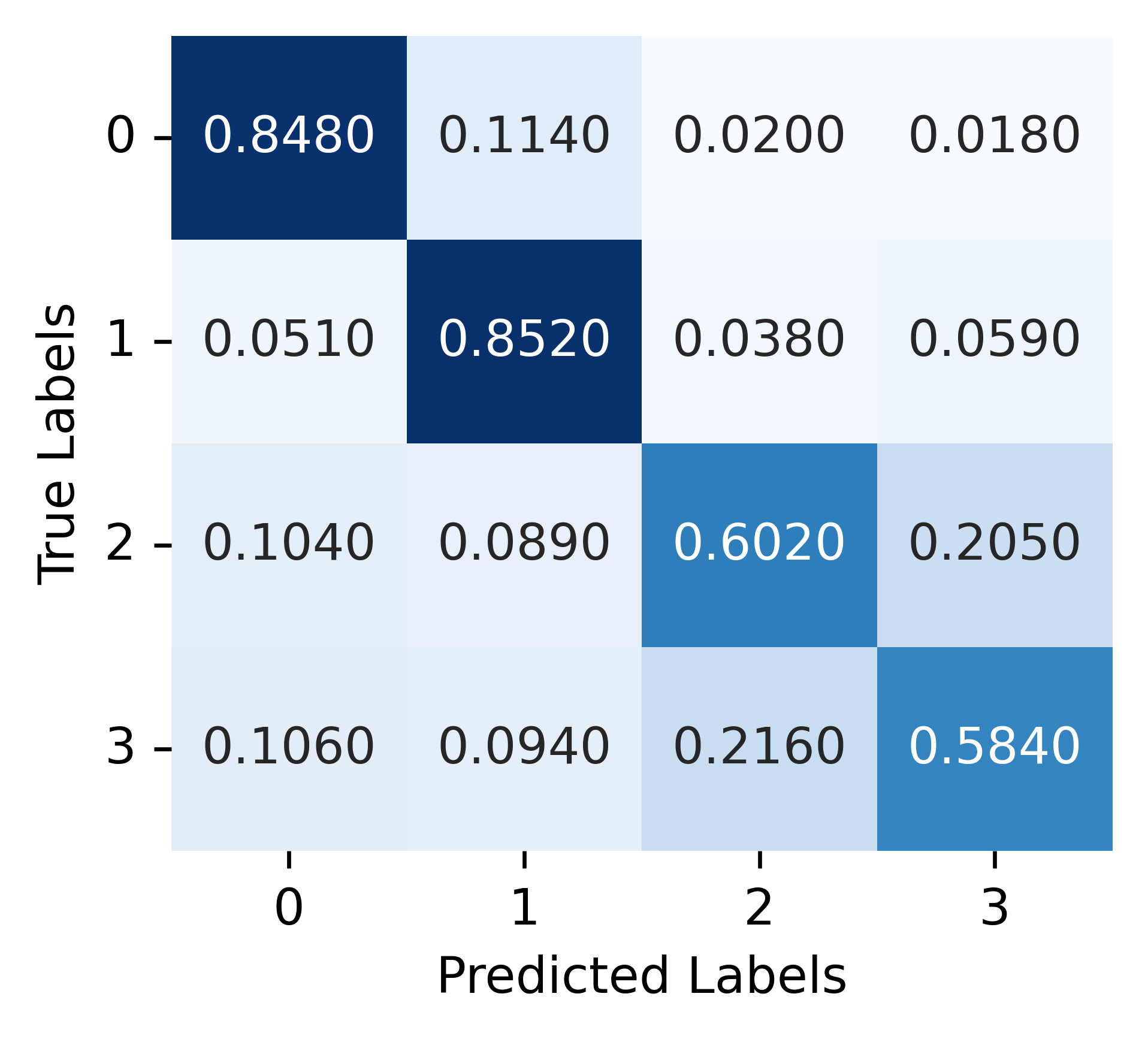}
    \end{minipage}%
    \begin{minipage}{0.47\columnwidth}
        \centering
        \includegraphics[width=\textwidth]{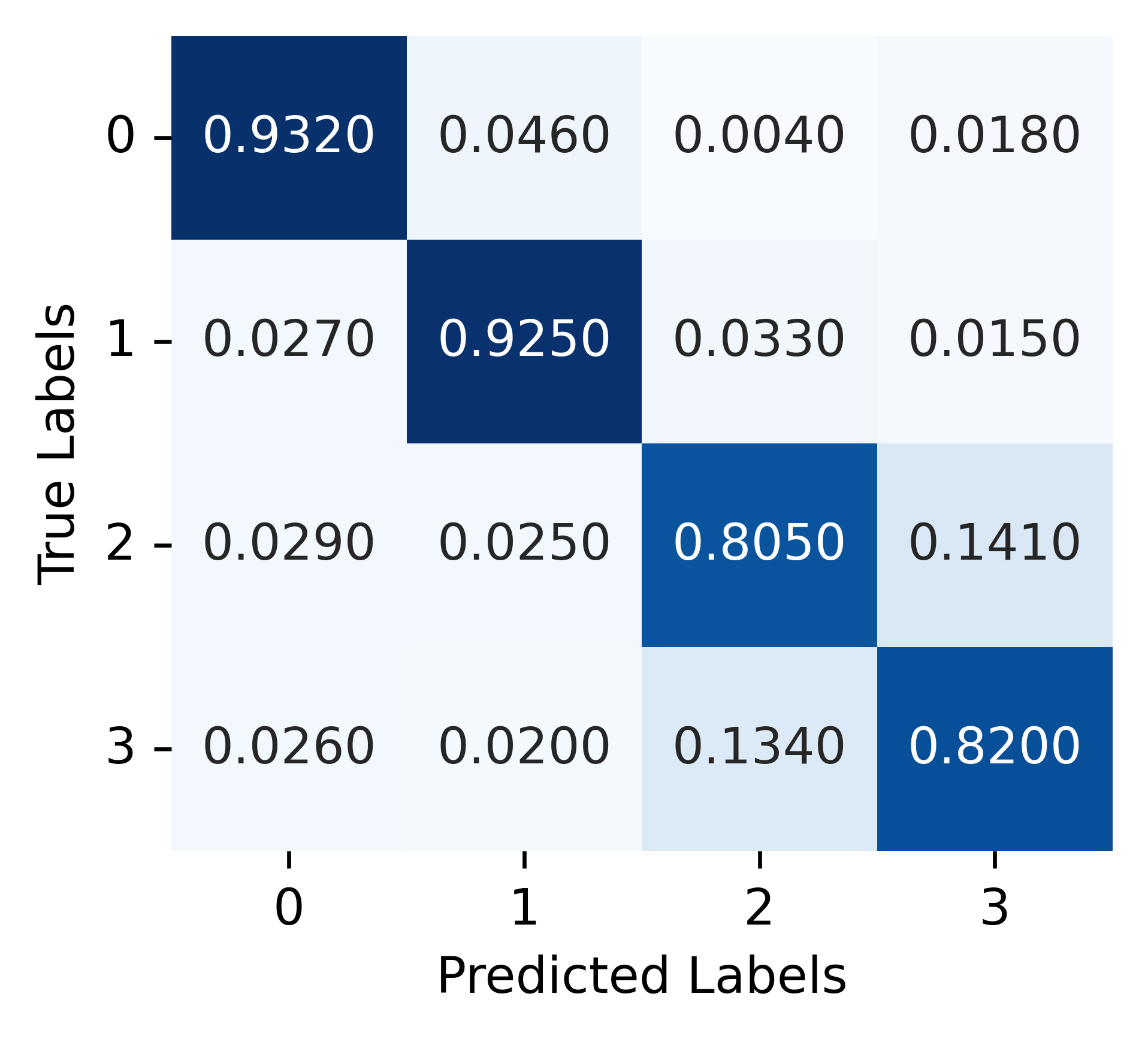}
    \end{minipage}

  \caption{Misclassification on actions between Assembly Station(class 2) and Manufacturing Station(class 3), located close to each other. Right: Ego-GRU, Left: Collective-GRU}
  \label{fig:confoosion}
\end{figure}

Additionally, a noticeable drop in performance was observed for the two stations placed in close proximity to each other, as shown in Figure~\ref{fig:confoosion}
% The consensus mechanism on top performs a vital task of keeping the overall decision of the multi-robot system constant. Moreover, the accuracy is also approximately equal to the collective-GRU values. 
\subsection{Effect of Number of Robots}
Figure~\ref{fig:num_robo} confirms the main advantage of our strategy: increasing the number of robots enhances the accuracy of action inference under the same conditions. For these experiments, we used Scenario 3 from our simulation setup (as described in Section~\ref{sec:sim}), observed data from the past 2 seconds, and predicted 2 seconds into the future while processing frames at 10Hz.

The results show a significant improvement in accuracy when increasing the number of robots from one to two, and from two to three. However, the improvement is less pronounced when adding a fourth robot. This diminishing return can be attributed to the observation that one of the four robots had a lower-quality view of the scene, resulting in reduced contributions to the collective perception. We hypothesize that further increasing the number of robots would enhance accuracy, as larger swarms tend to exhibit greater fault tolerance. In such cases, the impact of one or even several faulty robots is mitigated by the redundancy and diversity of functioning agents~\cite{milner2023swarm}.
%larger swarms 

\begin{figure}[ht]
    \centering
    \includegraphics[scale=0.87]{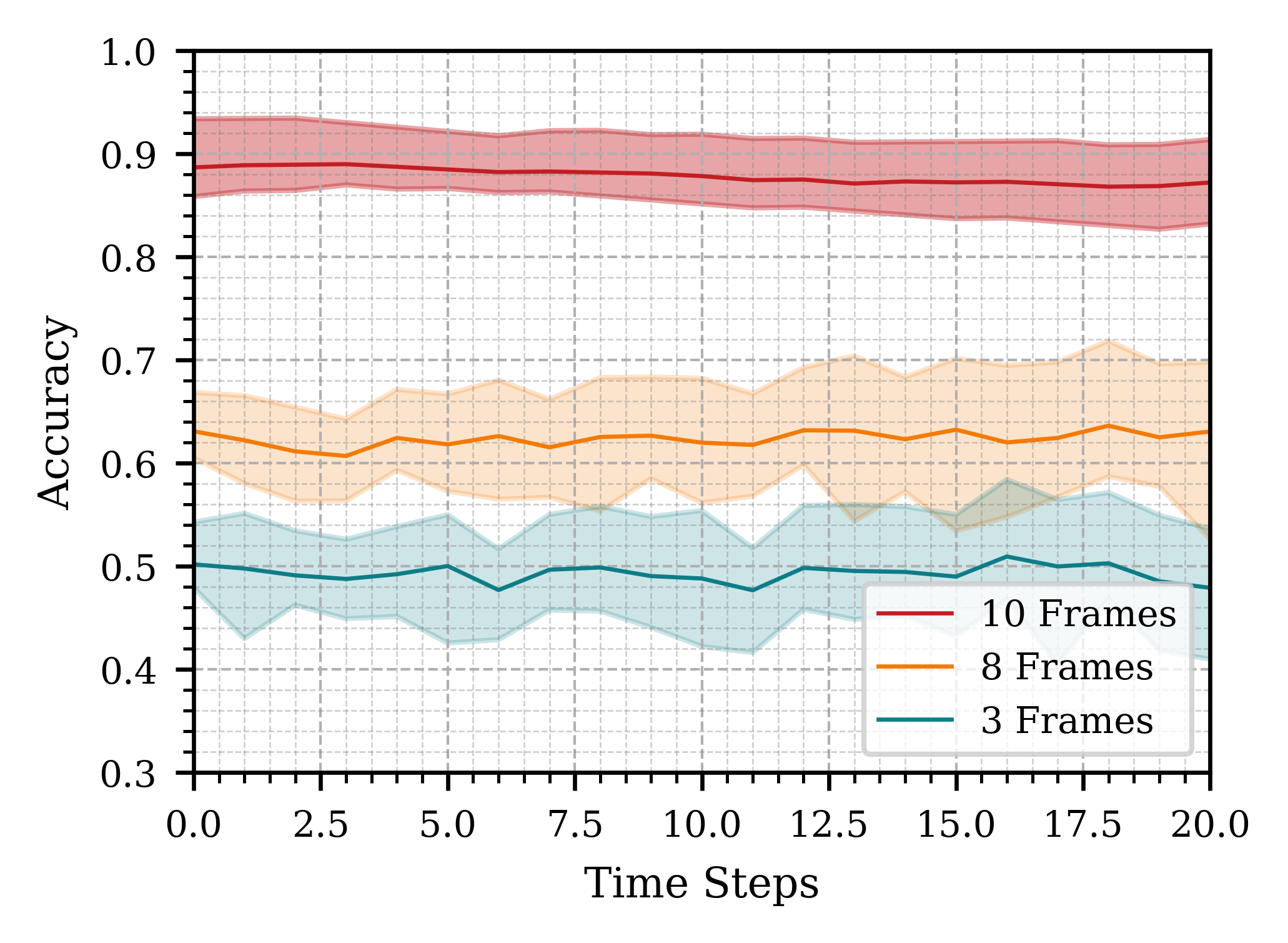}
    \caption{Effect of reducing the number of frames used for inference on prediction accuracy}
    \label{fig:frames}
\end{figure}

\subsection{Time Horizon Analysis}

We present the results for different time horizons in Figure~\ref{fig:horizon}, comparing the performance of using one robot versus three robots, with all data processed at 10 Hz. Overall, the results show that having more past information improves prediction accuracy. Specifically, predictions based on 2 seconds (20 frames) of past data outperform those based on 1 second (10 frames). For these cases, predictions extend 10 and 20 frames into the future, respectively, demonstrating better performance with increased historical data aggregation.

When the observation and prediction windows were extended to 3 seconds (30 frames), the system's performance became inconsistent. While the best-case predictions exceeded those of the 2-second case, the worst-case predictions were significantly worse. This inconsistency arises from the highly dynamic nature of the scene. Accumulating 30 frames increases the likelihood of encountering rapid movements or missed human detections, resulting in sequences of varying quality and more pronounced prediction fluctuations compared to shorter time horizons.

Additionally, processing longer time horizons required more computational resources and introduced greater latency. This highlights a trade-off between achieving higher potential accuracy and maintaining timely responsiveness, depending on the application's requirements. For our purposes, a 2-second time window offered the optimal balance between prediction accuracy and system responsiveness.

% \subsection{Impact of Frame Count} ### old
\subsection{Temporal resolution}
It is crucial to analyze how much temporal information is necessary to optimize computation and memory usage without compromising accuracy. We conducted additional experiments to compare the system's performance based on the number of frames used for prediction. Processing frames at 10 fps, we examined the effect on accuracy when using all 10 frames, 5 frames, and 3 frames. At runtime, we first accumulate all 10 frames and then pick the required number of frames at uniform intervals. Figure~\ref{fig:frames} illustrates the results.

Our observations indicate that utilizing all 10 frames leads to better predictions, highlighting the importance of sufficient temporal data for accurate action inference. 
% However, our codebase is designed with flexibility in mind, allowing it to adapt to different scenarios by handling variable input frequencies. This adaptability ensures that the system can be optimized for applications where computational resources or latency requirements necessitate processing fewer frames.
\begin{table}[htbp]
\caption{Performance Comparison of Models on Current Frame}
\label{tab:model_comparison}
\centering
\begin{tabular}{c l c}
\toprule
\textbf{Serial No.} & \textbf{Model} & \textbf{Avg on Frame $t$} \\
\midrule
% 1 & GNN & 68.27\% ± 10 \\
% 2 & GRU 1 Robot & 77.14\% ± 10 \\
% 3 & GRU Collective & 87.54\% ± 5 \\
% 4 & CVM & 60\% (Testing incomplete) \\
% 5 & 1D CNN & N/A \\
% 6 & Consensus & 89.12\% \\
1 & GNN & 70.73\% \\
2 & Ego-GRU & 77.7\% \\
3 & Collective-GRU & 88.87\% \\
4 & CVM & 66.1\%  \\
5 & 1D CNN & 63.31\% \\
6 & Collective-GRU + Consensus & 88.5\% \\
\bottomrule
\end{tabular}
\end{table}

\subsection{Comparison with Alternative Strategies}
Table~\ref{tab:model_comparison} presents the results from Scenario 3 of the simulation, using a 2-second temporal window, and compares them against the CVM and 1D-CNN strategies. While the CVM~\cite{scholler2020constant} performs well in Scenario 1, where no obstacles are present and the human moves along a straight path, its accuracy significantly declines in more complex scenarios. Specifically, the CVM struggles to predict human actions when the human changes direction, remains stationary, or stops.

In contrast, our spatiotemporal approach outperforms the 1D-CNN, demonstrating the advantage of integrating spatial and temporal information through GNNs and GRUs. The combination of spatial modeling via GNNs and temporal modeling through RNNs provides a comprehensive understanding of the scene, capturing both relationships between objects and the temporal evolution of actions. This holistic perspective enables our spatiotemporal pipeline to deliver more accurate predictions than methods relying solely on temporal sequences (such as the 1D-CNN) or simplistic motion assumptions (as in the CVM).

\subsection{Additional performance parameters}

\textbf{Quality of graphs} Robots with weaker or sparser graphs—where fewer objects in the scene are linked to the human node—tended to produce less accurate and more unpredictable predictions. The limited number of objects reduces the robot's ability to model relationships within the scene effectively, leading to decreased prediction performance.

Furthermore, individual robot predictions improved when the graph explicitly linked the source (the human’s current position) to the destination (the intended goal or station). This highlights the importance of graph structures that capture critical relationships between key entities in the environment, as these connections enhance the robot's spatial understanding.

\textbf{Erroneous data in small teams} In experiments with a relatively small multi-robot system of up to four robots, robots with weaker or sparser graphs not only made poor individual predictions but also negatively influenced the collective perception when their data was aggregated. This suggests that the quality of each robot's perception plays a significant role in the effectiveness of the consensus mechanism, particularly in smaller teams. These findings align with prior research~\cite{milner2023swarm}, underscoring the need for robust individual perception to ensure reliable collective decision-making in small multi-robot systems.

\section{Conclusion}
This paper presents a decentralized multi-robot shared perception pipeline that predicts the intent of a human operator in the scene by exploiting the spatial and temporal relationships between the human and surrounding objects. We demonstrated that having multiple robots observing the same scene from different viewpoints and sharing information can help develop a robust and scalable perception system.

% Currently, the system can handle only one human in the scene. Addressing scenarios with multiple humans will be a key focus in future developments. Additionally, our GNN currently utilizes Euclidean distances as edge attributes obtained from 2D images. To enhance the system's robustness, we aim to increase the dimensionality of the edge features to incorporate more contextual information, potentially using additional sensor modalities available on the robots.

% Currently, the system can handle only one human in the scene. Addressing scenarios with multiple humans will be a key focus in future developments. Additionally, our GNN currently utilizes Euclidean distances as edge attributes obtained from 2D images. To enhance the system's robustness, we aim to increase the dimensionality of the edge features to incorporate more contextual information, potentially using additional sensor modalities available on the robots. Furthermore, we plan to explore the integration of Vision-Language Models (VLMs) to improve scene understanding in dynamic and unfamiliar environments. Since the robots are mobile and there is a high probability that they will encounter new scenes they have not seen before, VLMs could enable richer semantic representations, thereby enhancing the robots' ability to interpret human actions and intentions in novel settings.
Currently, our system is limited to handling a single human in the scene, and expanding to multiple humans is a priority for future work. We also intend to enrich the GNN’s edge features—currently based only on Euclidean distances from 2D images—by incorporating higher-dimensional inputs derived from additional onboard sensors. Beyond that, we plan to integrate Vision-Language Models (VLMs) to enhance scene understanding in dynamic, unfamiliar environments. Such models could enable more robust semantic representations, improving the robot’s ability to interpret human actions and intentions in previously unseen scenarios.

Moreover, the consensus mechanism currently relies on the quantification of the robots' visibility but does not account for the quality of the view. Future work will focus on integrating view quality metrics into the consensus mechanism. We also plan to expand the action library to include more complex and realistic scenarios, closely resembling real-world industrial environments. Finally, we intend to implement and test our approach on real robots to validate its effectiveness in practical applications.

% \newpage
% \addtolength{\textheight}
\addtolength{\textheight}{-0.1cm}   % This command serves to balance the column lengths
                                  % on the last page of the document manually. It shortens
                                  % the textheight of the last page by a suitable amount.
                                  % This command does not take effect until the next page
                                  % so it should come on the page before the last. Make
                                  % sure that you do not shorten the textheight too much.
%%%%%%%%%%%%%%%%%%%%%%%%%%%%%%%%%%%%%%%%%%%%%%%%%%%%%%%%%%%%%%%%%%%%%%%%%%%%%%%%

%%%%%%%%%%%%%%%%%%%%%%%%%%%%%%%%%%%%%%%%%%%%%%%%%%%%%%%%%%%%%%%%%%%%%%%%%%%%%%%%

% \section*{ACKNOWLEDGMENTS}

%%%%%%%%%%%%%%%%%%%%%%%%%%%%%%%%%%%%%%%%%%%%%%%%%%%%%%%%%%%%%%%%%%%%%%%%%%%%%%%%

\bibliographystyle{IEEEtran}
\bibliography{references}

\end{document}